\documentclass[sigconf]{acmart}
\usepackage{hyperref}
\AtBeginDocument{%
  \providecommand\BibTeX{{%
    \normalfont B\kern-0.5em{\scshape i\kern-0.25em b}\kern-0.8em\TeX}}}

\copyrightyear{2021}
\acmYear{2021}
\setcopyright{licensedothergov}\acmConference[ASSETS '21]{The 23rd International ACM SIGACCESS Conference on Computers and Accessibility}{October 18--22, 2021}{Virtual Event, USA}
\acmBooktitle{The 23rd International ACM SIGACCESS Conference on Computers and Accessibility (ASSETS '21), October 18--22, 2021, Virtual Event, USA}
\acmPrice{15.00}
\acmDOI{10.1145/3441852.3471211}
\acmISBN{978-1-4503-8306-6/21/10}

\begin{document}

\title{Fluent: An AI Augmented Writing Tool for People who Stutter}


\author{Bhavya Ghai}
\affiliation{\institution{Stony Brook University}}
\email{bghai@cs.stonybrook.edu}

\author{Klaus Mueller}
\affiliation{\institution{Stony Brook University}}
\email{mueller@cs.stonybrook.edu}

\renewcommand{\shortauthors}{Ghai and Mueller, et al.}

\begin{abstract}
Stuttering is a speech disorder which impacts the personal and professional lives of millions of people worldwide. To save themselves from stigma and discrimination, people who stutter (PWS) may adopt different strategies to conceal their stuttering. One of the common strategies is word substitution where an individual avoids saying a word they might stutter on and use an alternative instead. This process itself can cause stress and add more burden. In this work, we present Fluent, an AI augmented writing tool which assists PWS in writing scripts which they can speak more fluently. Fluent embodies a novel active learning based method of identifying words an individual might struggle pronouncing. Such words are highlighted in the interface. On hovering over any such word, Fluent presents a set of alternative words which have similar meaning but are easier to speak. The user is free to accept or ignore these suggestions. Based on such user interaction (feedback), Fluent continuously evolves its classifier to better suit the personalized needs of each user. We evaluated our tool by measuring its ability to identify difficult words for 10 simulated users. We found that our tool can identify difficult words with a mean accuracy of over 80\% in under 20 interactions and it keeps improving with more feedback. 
Our tool can be beneficial for certain important life situations like giving a talk, presentation, etc. The source code for this tool has been made publicly accessible at \href{https://github.com/bhavyaghai/Fluent}{github.com/bhavyaghai/Fluent}.  
\end{abstract}


\maketitle

\section{Introduction}
Stuttering or Stammering is a speech disorder which is generally characterized by disfluencies like  
part-word repetitions eg. "I w-w-w-want a drink", prolonging a sound eg. "Ssssssssam is nice", blocks eg. "I want a (pause) cookie", etc.\cite{asha}. Possible causes of Stuttering include Family history, differences in how the brain works during speech, etc. According to the American Speech-Language-Hearing Association (ASHA), more than 70 million people worldwide stutter\cite{asha}. It is found in all parts of the world and impacts people across culture, race, sex, age, ethnicity, etc. \cite{guitar2013stuttering,andrews1983stuttering, zimmermann1983indians}. 

Stammering can have a profound negative impact on the personal and professional life of people who stutter (PWS). On the personal front, it can cause anxiety, frustration, embarrassment and public stigma\cite{prasse2008stuttering}. The general public has a less positive view of such people\cite{hulit1994association} and they might face discrimination and social devaluation\cite{boyle2018enacted}. On the professional front, such individuals might be perceived as less capable than their peers who do not stutter. For example, physicians who stutter are perceived to be more afraid, tense, nervous and to be less mature, intelligent and competent than their peers who do not stutter\cite{silverman1997nurses}. Similarly, lawyers are perceived as less educated, competent and intelligent than their peers who do not stutter \cite{silverman1990impact}. Also, 85\% of employers think that stuttering decreases employability; and only 9\% of employers think that people who stutter should be hired in a situation when two applicants are equally qualified \cite{klein2004impact}. Moreover, such individuals can also develop negative attitudes towards themselves \cite{boyle2018self}.


  

Speech therapy can be an effective tool for handling or living with stuttering. 
Multiple studies have corroborated the overall positive impact of stuttering treatment on an individual\cite{herder2006effectiveness,yaruss2010assessing}.
Having said that, there is no known cure for stuttering\cite{kalinowski2005stuttering}. Moreover, access to speech therapy can be limited\cite{verdon2011investigation} and it might be associated with blaming and shaming\cite{douglass2020speech}. 
As a result, many adults who stutter will continue to do so for their entire lives. 

Adults who stutter are likely to have been living with their stuttering since childhood. As a person gets older, they become more conscious of their situation and learn to anticipate stuttering moments better (also known as \textit{anticipation effect})\cite{anticipation}. 
When an individual anticipates stuttering, they use different strategies to hide their condition which are known as \textit{Avoidance behaviours}. This may include using fillers (e.g. ‘um', ‘like'), changing the feared word with its synonym (substitution), talk around the feared word (circumlocution), etc. Research has shown that avoidance behaviours are common among PWS\cite{murphy2007covert} and all people who stutter use such strategies to some degree\cite{starkweather1987fluency}. It should be noted that avoidance behaviours such as word substitution doesn't improve the underlying condition and can even have some negative effects\cite{boyle2016relations, boyle2018disclosure}. However, such tactics can help conceal stuttering with little or no observable disfluency. This might save a person from embarrassment, stress, etc. in professional settings like giving a talk, presentation, or in personal settings like dating, etc. \cite{constantino2017rethinking}.

\begin{figure*}
    \centering
  \includegraphics[width=\textwidth]{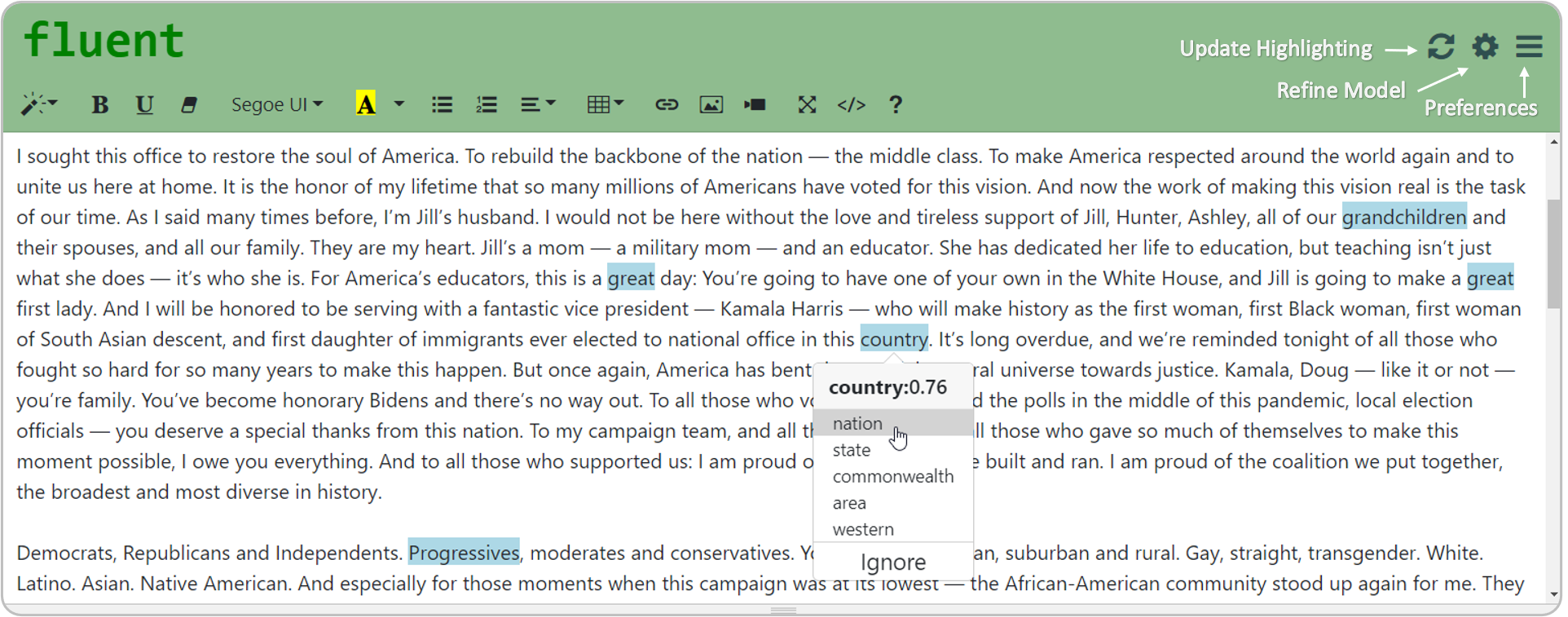}
  \caption{Visual Interface of Fluent. Words highlighted in blue are the ones which the user might find difficult to pronounce. Hovering over such words presents a set of alternatives (including Ignore option) which have similar meaning but might be easier to pronounce. In the above picture, the user hovers over the word ‘country' and the tool presents a set of alternatives namely, nation, state, commonwealth, area, etc. Buttons on the top right corner allows the user to provide explicit feedback (Refine Model) and provide a set of words which they find easy/difficult to pronounce (Preferences).}
  \label{fig:teaser}
  \Description[Visual interface of Fluent]{Fluent looks like any other editor with highlighted word which represent difficult to pronounce words. On hovering over a word, a popup appears which contains similar meaning words which can be easily pronounced. In the above picture, the user hovers over the word ‘country' and the tool presents a set of alternatives namely, nation, state, commonwealth, area, etc. Buttons on the top right corner allows the user to provide explicit feedback (Refine Model) and provide a set of words which they find easy/difficult to pronounce (Preferences).}
\end{figure*}

A recent study based on a large diverse set of PWS found that the goal of the majority of the participants (69.5\%) is to ‘not stutter' (hide stuttering) vs ‘stuttering openly' while speaking\cite{tichenor2019group}. One of the factors which may impact the likelihood of stuttering is phonological patterns\cite{howell2006phonetic}.   
In other words, PWS are likely to stutter on some words more than others. 
 To prevent oneself from stuttering, one has to identify which words they might struggle with and then think of a way to manage it. This process itself can take time, effort and can cause additional stress \cite{plexico2009coping}. In this work, we leverage recent advancements in AI such as phonetic embeddings to reduce this burden and help PWS hide their stuttering. 

We present \textit{Fluent}, a novel machine in  the loop writing tool for assisting PWS with writing scripts, which they can speak more fluently (minimize the number of stuttering events). Given a piece of text, our tool helps identify words that a person might struggle pronouncing (trigger words). Such words are highlighted in the user interface in real time. For each of such words, our tool provides a set of alternatives which have similar meaning but might be easier to pronounce. The user can simply hover over any of the highlighted words to choose from the set of alternatives (see \autoref{fig:teaser}). Since each user can have different requirements, our tool evolves itself via user feedback to provide better personalized support over time. 
Fluent can be used for writing speeches, scripts, dialogues which might be used by anchors, politicians, actors, etc.
This might not only save time but also enhance their confidence and ultimately impact their performance. 
The primary contributions of this work are as follows:-
\begin{itemize}
    \item We devise a novel method to identify words an individual might struggle pronouncing.
    \item We design and implement a new writing tool ‘Fluent’ which embodies our approach to identify trigger words and suggests suitable alternatives.
    \item We evaluate our system by measuring its ability to identify trigger words for 10 simulated users.
\end{itemize}


\section{Background and Related Work}



\subsection{AI for Stuttering}
Existing literature at the intersection of AI and stuttering focuses on building machine learning systems to identify and classify different types of disfluencies like Blocks, Prolongations\cite{esmaili2017automatic}, Sound Repetitions \cite{ravikumar2008automatic}, Interjections, etc. in speech utterances\cite{chee2009overview}. Such systems are typically trained on speech samples which are annotated for different kinds of disfluencies \cite{fluentnet,fassetti2019learning,swietlicka2013hierarchical,esmaili2017automatic,alharbi2020sequence}. Other approaches have leveraged data based on facial muscle movements\cite{das2020stuttering}, breathing patterns\cite{villegas2019novel}, brain activity\cite{brain_activity}, etc.      
The goal of such systems is to assist speech language pathologists (SLPs) during the stuttering assessment phase. During the assessment phase, the SLP counts the number and type of stuttering events in speech. This process can be tedious and subjective \cite{kully1988investigation}. Automated ML systems can perform this task and help save time while providing objective results.

Overall, this promising space seems under-explored as a lot of the work in this area pivots around a single problem i.e., stuttering detection. 
In this work, we have taken a different route of developing assistive writing technology for PWS by leveraging AI based methods. Unlike previous work, our system relies on self reported textual data to learn more granular phonetic patterns that an individual might struggle pronouncing. Moreover, our system utilizes user feedback to provide more personalized feedback. 



\begin{figure*}
 \centering 
 \includegraphics[width=0.95\textwidth]{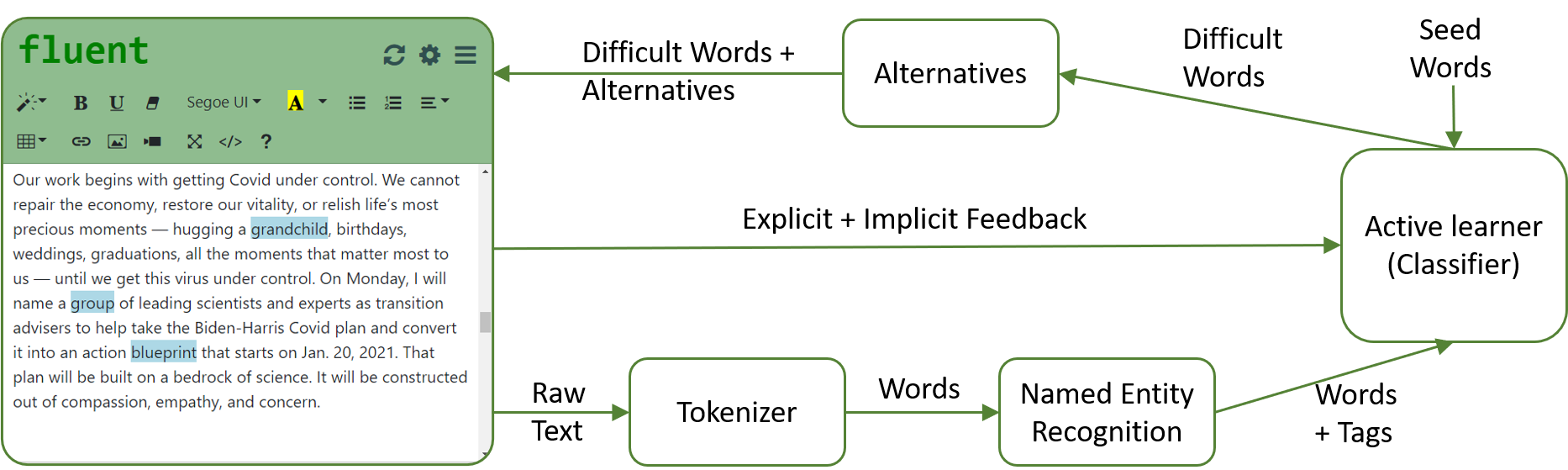}
 \caption{Workflow of Fluent}
 \label{fig:workflow}
 \Description[Workflow of Fluent]{This picture shows a flow diagram representing the workflow of the system. It includes components like Active learner, Visual Interface, Tokenizer, etc.}
\end{figure*}

\subsection{AI based Writing Tools}
The landscape of AI based writing tools typically
comprises of Natural Language Processing (NLP) based software systems aimed at improving productivity\cite{chen2012flow}, creativity\cite{frich2019mapping}, inclusion\cite{textio}, etc. \textit{Grammarly}\cite{karyuatry2018grammarly} is a popular AI writing assistant which provides real time suggestions for fixing grammatical errors, improved word choice, refining tone, etc. \textit{Textio}\cite{textio} is another AI powered writing tool which helps with hiring content. It suggests word level changes for writing inclusive and effective job descriptions. Similarly, there are other tools meant for a specialized task or audience. For example, \textit{FLOW}\cite{chen2012flow} is an interactive writing assistant for people who learn english as a foreign language, \textit{Creative Help}\cite{creative} and \textit{LISA}\cite{sanghrajka2017lisa} help with story writing, \textit{SWAN}\cite{SWAN} helps with scientific writing, etc.  

There has been some research on developing smart authoring tools for people with special needs. For example, Wu et al.\cite{wu2019design} developed a writing tool powered by Neural Machine Translation model to empower people with Dyslexia. There are also writing tools meant for people with visual impairments\cite{hanakovivc2006speech,waqar2019intelligent}, Sign Language Users\cite{ivanova2012bilingual}, etc.   
However, little to no attention has been paid towards people with speech disorders. 
In this paper, we have taken a small step to fill this gap by developing the first smart writing tool for people who stutter (PWS).

\subsection{Active Learning}
Supervised machine learning (ML) models require large amounts of labeled training data to provide good results. In many ML settings, unlabeled data points are abundant but labeling them can be time consuming and expensive\cite{zhu2005semi}. For such cases, Active learning (AL) serves as a viable learning paradigm as it focuses on training effective ML models using a minimum number of labeled training instances\cite{settles2009active}. 
AL achieves this by intelligently selecting data samples from a pool of unlabeled data which are then labeled by the oracle (e.g., a human annotator)\cite{unc}. The ML model is retrained iteratively for every new set of labeled data points as they come in. This process continues for a predefined number of iterations or until the annotation budget lasts\cite{stopping}. Here, the data points can be selected using different sampling strategies 
like Uncertainty sampling\cite{unc}, Query by committee sampling\cite{qbc,qbc2}, Hierarchical sampling\cite{dasgupta2008hierarchical}, QUIRE\cite{quire}, etc. Active learning has been found to work well for different applications\cite{ramirez2017active,xal} including Text classification\cite{tong2001support}, Named entity recognition\cite{chen2015study}, etc. However, it comes with its own set of practical challenges\cite{kagy2019practical,settles2011theories}. Research has shown that its benefits might not generalize reliably across models and tasks\cite{lowell2018practical}. In this work, we are trying to leverage AL in the context of stuttering. More specifically, we are investigating if AL can help learn the unique phonetic patterns that an individual might struggle pronouncing.    

\section{Fluent}

\subsection{Design Goals}
We have identified the following design goals based on the existing literature on stuttering, interactive systems\cite{wright2021recast} and the personal experiences of an author of this paper who stutters:

\begin{itemize}

    \item [\textbf{G1.}] \textbf{Identify Difficult Words}:
Given a piece of text, our tool should identify words which a given individual might find hard to pronounce. Here, each individual might struggle with different sounds, syllables, etc. Our goal is to build a generic tool which can learn the individual requirements of each user to provide personalized support. 
    
    \item [\textbf{G2.}] \textbf{Alternatives}: Given a word which might be hard to pronounce, our tool should provide a set of alternatives which have similar meaning and can can be easily pronounced by the individual it is targeted to. 

     \item [\textbf{G3.}] \textbf{Interface Design}: The Interface should be designed such that it is easy to latch onto (intuitive), accessible and should shield the user from underlying technical details (minimalist). It should provide the desired functionality while ensuring minimal lags to ensure smooth user experience. 
     
\end{itemize}

\subsection{Identifying Difficult to Pronounce Words}
An intuitive solution to classify words based on their pronunciation can be to use acoustic word embeddings. Such embeddings provide a vector representation for a given speech signal. However, we might get different embeddings for the same word pronounced by people from different gender, age, etc.~\cite{van2020improving}. Moreover, dealing with sound can be more computationally expensive (defeating G3). Hence, we used Phonetic embeddings~\cite{parrish2017poetic} based on the CMU Pronouncing Dictionary which is independent of speaker bias. Phonetic embeddings map each word to its corresponding vector representation based on the constituting phonemes. Words with similar pronunciation will be closer to each other in the embedding space.   
\begin{figure}
  \centering
   \includegraphics[scale=0.30]{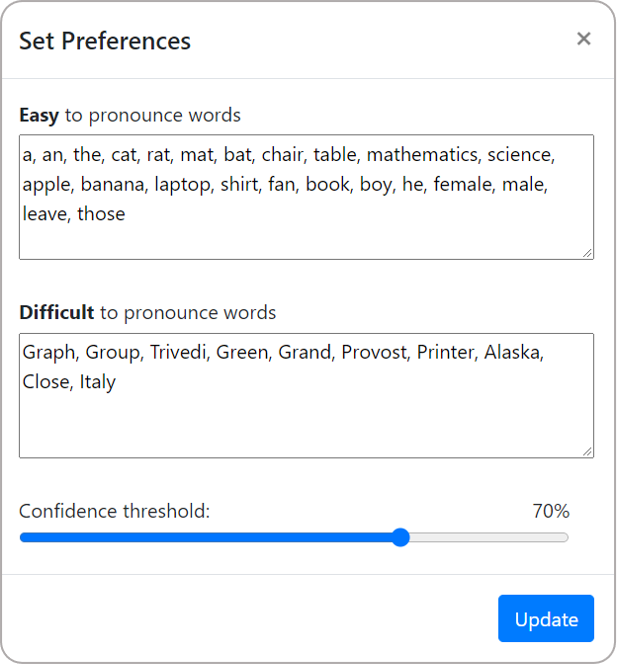}
  \captionof{figure}{User Preferences. The user can provide details on which words they find easy/difficult to pronounce.}
  \label{fig:prefer}
  \Description[Dialog box for User Preferences]{The picture contains a pair of text boxes where the user is expected to list the words they find easy/difficult to pronounce.}
\end{figure}

In the beginning, the user is asked to list a set of at least 5 words which they find easy and difficult to pronounce, respectively (see Fig. \ref{fig:prefer}). The larger the number of words, the better. Such words are also referred as seed words (see \autoref{fig:workflow}). Here, we are expecting the user to have a certain level of self awareness about their condition. Each word in either lists is mapped to its corresponding phonetic embedding\cite{parrish2017poetic}. Thereafter, we train an off the shelf binary SVM model (Active learner) to classify between easy and difficult words. For a given word, this trained model returns a numerical score between 0 and 1 which represents the likelihood that the word is hard to pronounce. We use this classifier over each word in the text editor and highlight words whose probability of being difficult is greater than a threshold. By default, the threshold value is 0.7. The user is free to change this threshold. A higher threshold might result in more false negatives and a lower threshold might lead to more false positives. Here, we have used a spacy\footnote{https://spacy.io/ \label{spacy}} based tokenizer to extract words from the raw text in the interface.  


\subsection{Adapting to Personalized needs}
The initial classification model trained over few training instances can be suboptimal. 
To refine the model further and better adapt it to personalized needs, we utilize explicit and implicit feedback from the user. 
Implicit feedback is gathered based on which option is selected from the dropdown menu when hovering over a highlighted word. If the user chooses the ‘Ignore' option, we add the highlighted word to our list of \textit{easy} words. Else we add it to the list of \textit{hard} words and add the alternative word chosen to the \textit{easy} list. Thereafter, we retrain the model over the updated word lists.   

\begin{figure}
  \centering
  \includegraphics[scale=0.35]{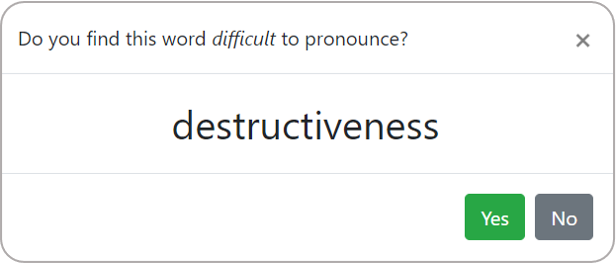}
  \captionof{figure}{Explicit Feedback: Query for refining Active learning classifier}
  \label{fig:query}
  \Description[Explicit Feedback]{Picture shows a popup which asks a question i.e., Is the word ‘chemical' difficult to pronounce?}
\end{figure}

For explicit feedback, we ask the user to indicate if a specific word is difficult or not (see \autoref{fig:query}). Here, the chosen word is selected from the pool of all unlabeled words in the phonetic embedding about which the current SVM model is most uncertain about. More specifically, we use entropy based uncertainty sampling\cite{settles2009active} to select the next word. Based on the user provided label, the word is added to an existing list of \textit{easy} or \textit{hard} words. Thereafter, the model is retrained based on the updated word lists and the next word with highest predictive uncertainty is chosen.
This iterative process might lead to fast convergence of the SVM model while requiring minimum number of labeled instances. 

\subsection{Finding Alternative Words}
To find suitable alternatives for a given word, we first generate a list of words which have similar meaning. Thereafter, we discard those synonyms which might be difficult to say as per the current SVM classifier. One possible option to generate synonyms is using online dictionaries like Thesaurus. Such dictionaries provide high quality synonyms but are limited in their coverage i.e., they do not provide synonyms for many words. Another possible approach is to use nearest neighbors as synonyms in word embeddings like word2vec\cite{word2vec}. This approach provides broad coverage but the quality of synonyms suffer. In this work, we have used the DataMuse API\footnote{https://www.datamuse.com/api/} which uses multiple online dictionaries, WordNet, Word2Vec and other databases to yield good quality synonyms for most words. Moreover, we identify words which represent names, places, dates, etc. using Named Entity Recognition from the spacy$^{\ref{spacy}}$ package.
Our tool just highlights such words and leaves it to the user to deal with such cases; it doesn't generate alternatives for such words.  

\subsection{Interface Design}
Fluent is implemented as a web application using python based web framework \textit{Flask}. The visual interface (see Fig.\ref{fig:teaser}) is built using javascript based open source library \textit{Summernote}. This makes Fluent easily accessible using a web browser across different platforms without needing any third party software or specialized computing resources. The default visual interface is designed to look like a generic rich text editor. Additional features pertaining to stuttering popup only upon clicking buttons on the top right of the interface. This is to ensure easy flow of ideas without getting distracted by too many options. For highlighting a word, we took inspiration from popular writing tools like Textio\cite{textio} which also uses a different background color to highlight a word. To display the set of alternative words, we have used a popup mechanism populated with alternatives which appears right below the word being hovered over.  
Such a design choice is implemented to mimic spell checkers and other tools like Grammarly\cite{karyuatry2018grammarly} which most people might have interacted with at some point. This might help users to quickly latch on to our interface without needing additional training.

\begin{figure*}
\centering
\begin{minipage}{.33\textwidth}
  \centering
    \includegraphics[scale=0.4]{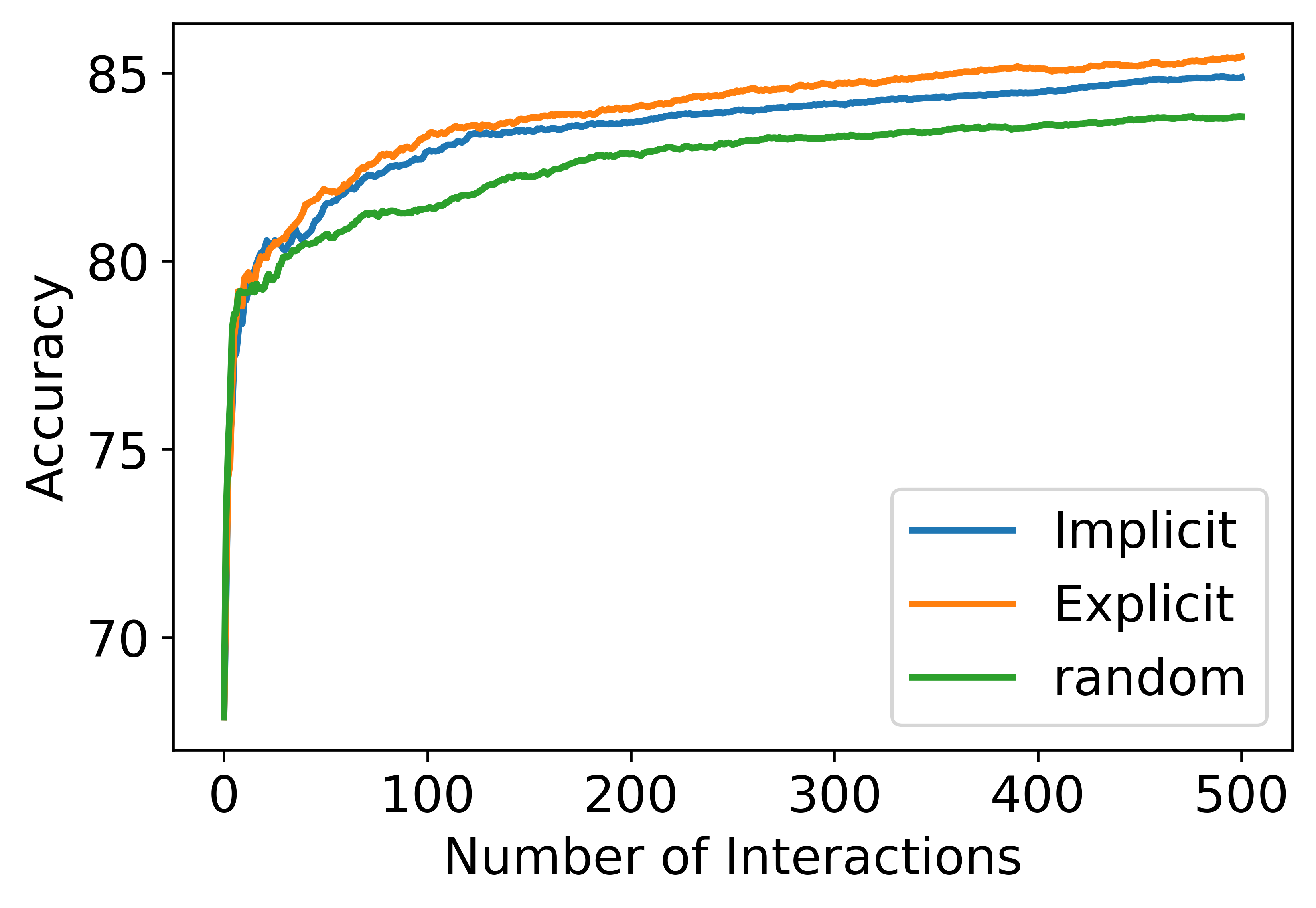}
  \label{fig:acc}
\end{minipage}%
\begin{minipage}{.33\textwidth}
  \centering
  \includegraphics[scale=0.4]{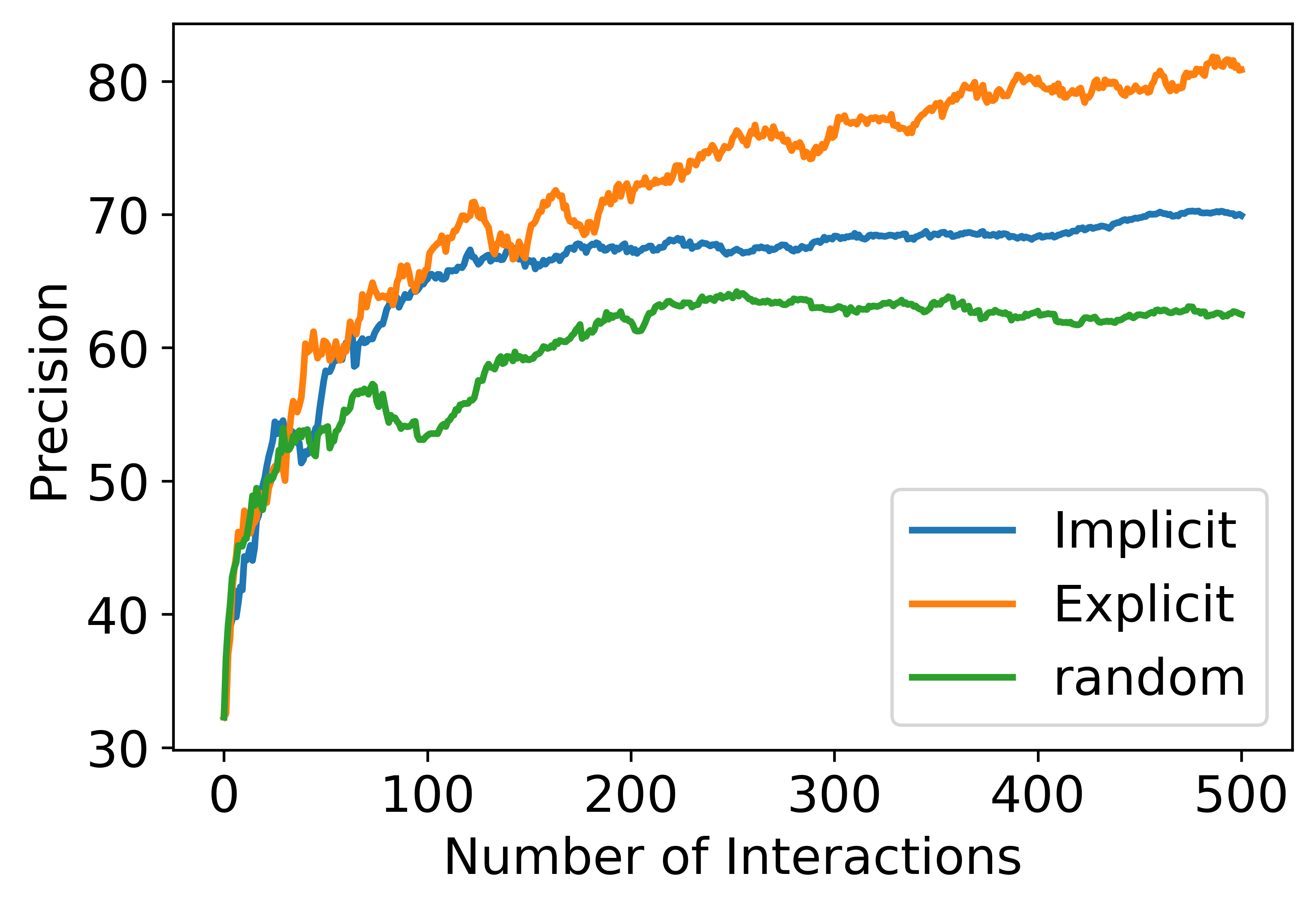}
  \label{fig:pre}
\end{minipage}
\begin{minipage}{.33\textwidth}
  \centering
  \includegraphics[scale=0.4]{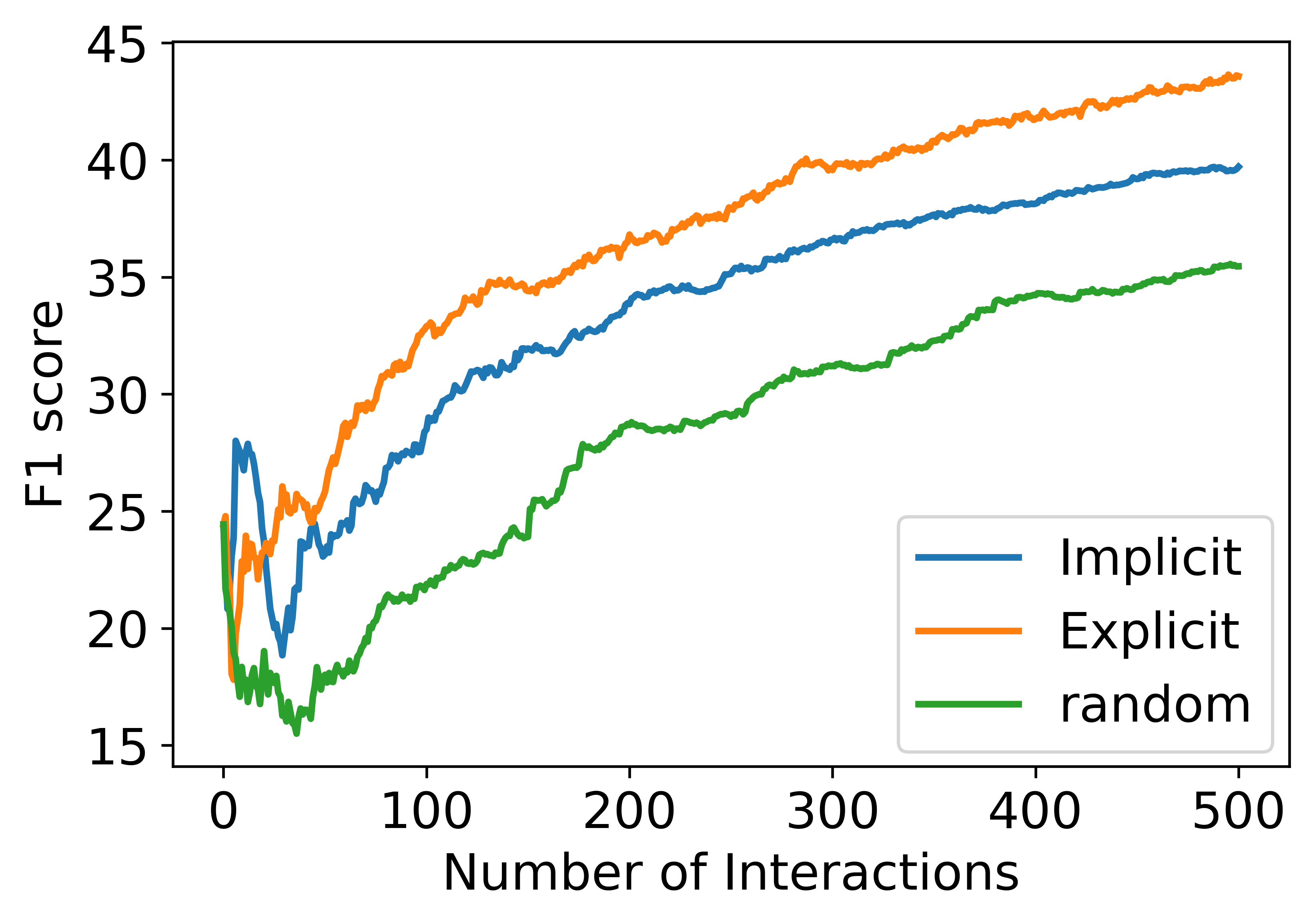}
  \label{fig:f1}
\end{minipage}
\vspace{-1em}
\caption{Plots showing the mean values for Accuracy, Precision \& F1 Score for 10 simulated users across 500 interactions.}
\label{fig:eval}
\Description[Evaluation metrics]{The picture shows 3 line charts which represents the mean values for Accuracy, Precision \& F1 Score for 10 simulated users across 500 interactions. All the plots show a positive trend. }
\end{figure*}

\section{Evaluation}
We evaluate our system in term of its ability to identify trigger words for 10 hypothetical users who interact with the system in a specific manner. 
\paragraph{User Profiles} We intended to evaluate our system on a diverse set of realistic user profiles with varying degrees of stuttering. So, we modeled 4 different user profiles based on self reported data from stuttering communities on facebook/reddit and one from the personal experiences of an author of this paper who stutters. Here, each user profile is defined by a phonetic pattern that they find difficult to pronounce and their corresponding seed words (see Supplementary material).  
For example, User 1 struggles to pronounce words which start with a consonant followed by a ‘r' sound.
To model severe stuttering for the last 5 profiles, we used a combination of two or three phonetic patterns from the top 5 profiles. For example, User 10 struggles with 3 phonetic patterns i.e., words starting with ‘st' or ‘fl', words with an ‘r' or ‘l' letter at the second place and words with a ‘sc' or ‘ch' sound anywhere in the word.

\paragraph{Data} We used 2,467 TED talks transcripts\footnote{Dataset link: https://www.kaggle.com/rounakbanik/ted-talks?select=transcripts.csv} data which contains ~57k unique words to evaluate our tool. 
We computed the ground truth (true label) for each word across all users based on their respective phonetic pattern.
For each unique word, we got 10 binary labels corresponding to each user. With the ground truth available, we split the 57k words into train and test data in the ration 75:25. We evaluate our classifier on how well it can predict the true label for the test data.

\paragraph{Simulation} At the beginning, each user provides 5 easy and 5 difficult words based on their unique condition. We train a SVM classifier based on these words. 
Thereafter, we have simulated two scenarios where the user provides only explicit or implicit feedback. This is to investigate how well each feedback mechanism works individually and relative to each other. In reality, a user can choose to provide both forms of feedback. For reference, we have also added a third scenario where we label words randomly. To simulate each interaction for the implicit feedback scenario (confidence threshold = 0.1.), each user searches for the first highlighted word among the TED dataset and interacts with it. Here, the user chooses the ‘Ignore' option if the highlighted word is a false positive. Otherwise, the user chooses the first alternative suggested by the system. 
For explicit feedback, we assume that each user will provide the correct response for all queries. After each explicit/implicit feedback (interaction), word lists are updated and the model is retrained. Thereafter, we measure metrics like precision, accuracy and f1 score corresponding to each user over the test dataset. Higher values for such metrics indicates better personalized support. 


\paragraph{Results} \autoref{fig:eval} shows the mean scores for accuracy, precision and F1 score for 10 users across 500 interactions. We can see an overall positive trend for both forms of feedback across different metrics. In under 20 interactions, the classifier reached a mean accuracy of over 80\% for both forms of feedback (random classifier will yield 50\%). This shows that both forms of feedback are effective in enhancing the performance of our classifier. It is interesting to observe that explicit feedback is more effective than implicit feedback and random labeling across different metrics. It should be noted that an implicit feedback can potentially add two data points (if the highlighted was actually a trigger word) compared to a single data point for explicit feedback per interaction. This suggests that active learning can significantly accelerate the learning process in the context of stuttering. So, it is advisable that the user provides explicit feedback whenever possible to accelerate the learning process.

\begin{figure}
  \centering
   \includegraphics[scale=0.40]{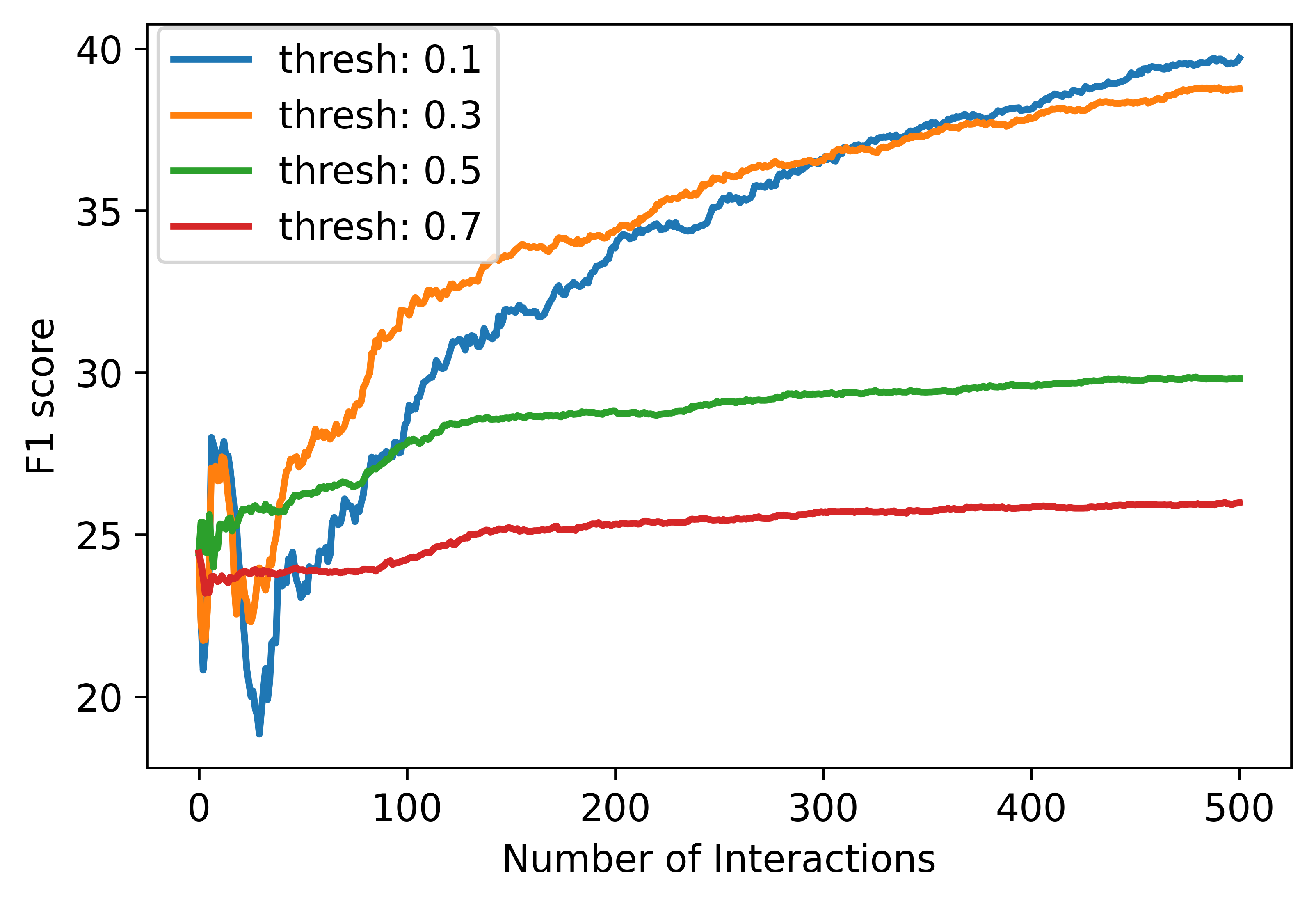}
  \captionof{figure}{Plot showing how F1 score varies for different thresholds. Larger the threshold, lower the mean F1 score. }
  \label{fig:implicit}
  \Description[Line chart for different threshold for Implicit feedback]{The picture contains 4 different lines which shows F1 scores across number of interaction. A Lower threshold has higher F1 score. }
\end{figure}

Implicit feedback might be a bit slower but it may be a more natural and non-intrusive way to provide feedback. To help users choose a threshold wisely (see \autoref{fig:prefer}), we explored the relation between confidence threshold and F1 score for the implicit feedback scenario (see \autoref{fig:implicit}). Our experiment suggests that F1 score can vary significantly for different threshold values and a lower threshold might generally yield higher F1 score.

Overall, our experiments demonstrate that our approach of using phonetic embeddings combined with active learning provides promising results. Our tool can effectively learn the personalized needs of different users within a short span of time.

\section{Discussion, Limitations and Future Work}

\paragraph{Intended Use} Fluent is designed to assist PWS get through certain important life situations like giving a talk. In the process, it might make the person more self-aware about their condition by helping them discover trigger words which they were previously unaware of. It might also enrich their vocabulary by suggesting alternatives. However, it should be noted that word substitution is a just a coping mechanism and it doesn't improve the underlying condition. Our tool just helps at concealing the underlying condition better. Avoiding behaviors can be empowering\cite{constantino2017rethinking} but they can also have some negative effects. For example, research has shown that people who try to conceal stuttering report lower levels of self-esteem and quality of life\cite{boyle2016relations, boyle2018disclosure}. Users are advised to visit a speech language pathologist for proper personalized treatment. 

\paragraph{Design} This work ventures into a previously unexplored territory of building a smart text editor for PWS. So, there are no existing design guidelines to build such a system. In this work, we have emulated the general design principles of smart text editors like Grammarly and relied upon the personal experiences of an author of this paper who stutters to make certain design choices. Such design decisions may/mayn't generalize to the PWS at large. Future work might involve PWS to evaluate the design decisions made in this work and devise a set of comprehensive design guidelines to inform the development of such tools in the future. 

\paragraph{Efficacy}
We have focused on evaluating the learning capability of our system from a machine learning perspective. Our promising results are based on 10 simulated users who interacted with the system in a specific predefined fashion. 
In the real world, people might struggle on varied sets of words and and might interact with the system in different ways. 
The next step will be to perform a comprehensive evaluation from a human centered perspective. It will be interesting to conduct an empirical study with PWS and investigate the usability, utility and effectiveness of Fluent relative to a plain text editor. Future work might explore the utility of this tool for speech therapy and also measure the impact of this tool on an individual's confidence and stress levels while speaking. 

\paragraph{Substitution} Each word in the English language has its own subtle meaning. So, it might not be possible to substitute a word completely. However, we have tried to suggest similar meaning words by leveraging knowledge from multiple dictionaries, WordNet\cite{wordnet}, word2vec embedding\cite{word2vec}, etc. Future work might incorporate contextualized word embeddings like BERT\cite{bert}, etc. to further refine the set of alternatives for a given word. It should be noted that words representing immutable categories like names, places, dates, etc. can't be substituted. Our tool just highlights such cases and leaves it to the user to deal with them.
  
 
\paragraph{Target Users}  
Fluent requires the user to provide seed words in the beginning and assumes that the subsequent user feedback is mostly accurate. This requires the user to have a certain level of self awareness about one's condition. Moreover, if one suffers from acute stammering, then one might need to substitute multiple words in a sentence which might break the semantic structure of the sentence. So, our tool might not work well for such people. Future work might explore the utility of our tool for non-native speakers and other speech disorders like lisp. For example, people who have a lisp might find certain sounds like "s" hard to pronounce. So, they might struggle on words like misses, session, aesthetics, etc. Similarly, native french speakers find words like thorough, through, clothes, etc. difficult because ‘th’ sound doesn’t exist in french. It will be interesting to see how well our tool can identify words with such underlying phonetic pattern.  

\paragraph{Vocabulary} Fluent can analyze and possibly highlight only those words which are present in the phonetic embedding. In our case, the phonetic embedding has more than 116k unique words which is substantially larger than an average person's active vocabulary size of \textasciitilde20k words. So, our tool should be able to deal with most words used by an average person. However, it might not contain domain specific jargon. Moreover, our tool can't deal with numbers say 64, abbreviations say NY and symbols say \$ unless they are written in worded form like Sixty four, New York, etc. Future work might might try to deal with such cases and extend support for other languages (not just english).  



\paragraph{Granularity} Fluent processes running text by breaking it down into words and then analyzes each one of them individually. 
It is possible that an individual might pronounce a word (say ‘juror') well on its own but might struggle when the same word occurs in a specific context/sentence (say ‘rural juror'). Our current system doesn't capture such cases as it analyzes each word on its own without considering its context. Future work might account for such cases by analyzing word n-grams (contiguous sequence of words) apart from just individual words.    

\paragraph{User Feedback} Fluent elicits binary user feedback i.e., whether a word is difficult to pronounce or not. Such a feedback is less cognitively taxing on the user and can also be gathered in a non-intrusive fashion (Implicit feedback). However, the real situation might be more subtle. For e.g., a person might struggle on a word sometimes but not always. To cater to such situations, our tool can be easily modified to elicit to a more nuanced (continuous) feedback instead of a discrete binary label. For eg., we can ask the user to rate their confidence in the binary label or to rate the difficulty of a word on a n-point likert scale. Such nuanced feedback can be incorporated by existing active learning frameworks for more effective learning\cite{song2018active}. Here, we should also study and consider the kinds of feedback PWS naturally want to provide\cite{xal}. 
Apart from severity, stuttering behavior can also vary with time and situation\cite{variability}. Future work should also try to capture such variability.


\section{Conclusion}
We have presented \textit{Fluent}, a smart text editor meant for people who stutter. Fluent embodies a novel adaptive, interactive and iterative technique to identify words which an individual might struggle pronouncing and proposes suitable alternatives. Our experiments show promising results and corroborate this problem as a possible research direction. We hope this work will encourage other researchers to work on this important and under-explored area. In the long run, we hope our work might inspire popular tools like MS PowerPoint, MS Word, etc. to add accessibility features for PWS.
\begin{acks}
We want to thank Xiaojun Bi (Stony Brook University), Yao Du (Monmouth University), Erik X. Raj (Monmouth University) and Renee Fabus (Stony Brook University) for their support and guidance. We would also like to thank the anonymous reviewers for their feedback. This research was supported by NSF grant IIS 1941613. 
\end{acks}

\bibliographystyle{ACM-Reference-Format}
\bibliography{main}


\begin{thebibliography}{70}


\ifx \showCODEN    \undefined \def \showCODEN     #1{\unskip}     \fi
\ifx \showDOI      \undefined \def \showDOI       #1{#1}\fi
\ifx \showISBNx    \undefined \def \showISBNx     #1{\unskip}     \fi
\ifx \showISBNxiii \undefined \def \showISBNxiii  #1{\unskip}     \fi
\ifx \showISSN     \undefined \def \showISSN      #1{\unskip}     \fi
\ifx \showLCCN     \undefined \def \showLCCN      #1{\unskip}     \fi
\ifx \shownote     \undefined \def \shownote      #1{#1}          \fi
\ifx \showarticletitle \undefined \def \showarticletitle #1{#1}   \fi
\ifx \showURL      \undefined \def \showURL       {\relax}        \fi
\providecommand\bibfield[2]{#2}
\providecommand\bibinfo[2]{#2}
\providecommand\natexlab[1]{#1}
\providecommand\showeprint[2][]{arXiv:#2}

\bibitem[\protect\citeauthoryear{??}{ash}{2020}]%
        {asha}
 \bibinfo{year}{Accessed August 2020}\natexlab{}.
\newblock \bibinfo{title}{Stuttering}.  (\bibinfo{year}{Accessed August 2020}).
\newblock
\urldef\tempurl%
\url{https://www.asha.org/stuttering/}
\showURL{%
\tempurl}
\newblock
\shownote{ASHA: American Speech-Language-Hearing Association.}


\bibitem[\protect\citeauthoryear{Alharbi, Hasan, Simons, Brumfitt, and
  Green}{Alharbi et~al\mbox{.}}{2020}]%
        {alharbi2020sequence}
\bibfield{author}{\bibinfo{person}{Sadeen Alharbi}, \bibinfo{person}{Madina
  Hasan}, \bibinfo{person}{Anthony~JH Simons}, \bibinfo{person}{Shelagh
  Brumfitt}, {and} \bibinfo{person}{Phil Green}.}
  \bibinfo{year}{2020}\natexlab{}.
\newblock \showarticletitle{Sequence labeling to detect stuttering events in
  read speech}.
\newblock \bibinfo{journal}{\emph{Computer Speech \& Language}}
  \bibinfo{volume}{62} (\bibinfo{year}{2020}), \bibinfo{pages}{101052}.
\newblock


\bibitem[\protect\citeauthoryear{Altschuler and Bloodgood}{Altschuler and
  Bloodgood}{2019}]%
        {stopping}
\bibfield{author}{\bibinfo{person}{Michael Altschuler} {and}
  \bibinfo{person}{Michael Bloodgood}.} \bibinfo{year}{2019}\natexlab{}.
\newblock \showarticletitle{Stopping active learning based on predicted change
  of f measure for text classification}. In \bibinfo{booktitle}{\emph{2019 IEEE
  13th International Conference on Semantic Computing (ICSC)}}. IEEE,
  \bibinfo{pages}{47--54}.
\newblock


\bibitem[\protect\citeauthoryear{Andrews, Hoddinott, Craig, Howie, Feyer, and
  Neilson}{Andrews et~al\mbox{.}}{1983}]%
        {andrews1983stuttering}
\bibfield{author}{\bibinfo{person}{Gavin Andrews}, \bibinfo{person}{Susan
  Hoddinott}, \bibinfo{person}{Ashley Craig}, \bibinfo{person}{Pauline Howie},
  \bibinfo{person}{Anne-Marie Feyer}, {and} \bibinfo{person}{Megan Neilson}.}
  \bibinfo{year}{1983}\natexlab{}.
\newblock \showarticletitle{Stuttering: A review of research findings and
  theories circa 1982}.
\newblock \bibinfo{journal}{\emph{Journal of speech and hearing disorders}}
  \bibinfo{volume}{48}, \bibinfo{number}{3} (\bibinfo{year}{1983}),
  \bibinfo{pages}{226--246}.
\newblock


\bibitem[\protect\citeauthoryear{Boyle}{Boyle}{2016}]%
        {boyle2016relations}
\bibfield{author}{\bibinfo{person}{MP Boyle}.} \bibinfo{year}{2016}\natexlab{}.
\newblock \showarticletitle{Relations between stuttering disclosure and
  self-empowerment in adults who stutter}. In \bibinfo{booktitle}{\emph{Poster
  presented at the American Speech-Language-Hearing Association, Annual
  Convention, Philadelphia, PA}}.
\newblock


\bibitem[\protect\citeauthoryear{Boyle}{Boyle}{2018}]%
        {boyle2018enacted}
\bibfield{author}{\bibinfo{person}{Michael~P Boyle}.}
  \bibinfo{year}{2018}\natexlab{}.
\newblock \showarticletitle{Enacted stigma and felt stigma experienced by
  adults who stutter}.
\newblock \bibinfo{journal}{\emph{Journal of Communication Disorders}}
  \bibinfo{volume}{73} (\bibinfo{year}{2018}), \bibinfo{pages}{50--61}.
\newblock


\bibitem[\protect\citeauthoryear{Boyle and Fearon}{Boyle and Fearon}{2018}]%
        {boyle2018self}
\bibfield{author}{\bibinfo{person}{Michael~P Boyle} {and}
  \bibinfo{person}{Alison~N Fearon}.} \bibinfo{year}{2018}\natexlab{}.
\newblock \showarticletitle{Self-stigma and its associations with stress,
  physical health, and health care satisfaction in adults who stutter}.
\newblock \bibinfo{journal}{\emph{Journal of Fluency Disorders}}
  \bibinfo{volume}{56} (\bibinfo{year}{2018}), \bibinfo{pages}{112--121}.
\newblock


\bibitem[\protect\citeauthoryear{Boyle, Milewski, and Beita-Ell}{Boyle
  et~al\mbox{.}}{2018}]%
        {boyle2018disclosure}
\bibfield{author}{\bibinfo{person}{Michael~P Boyle}, \bibinfo{person}{Kathryn~M
  Milewski}, {and} \bibinfo{person}{Carolina Beita-Ell}.}
  \bibinfo{year}{2018}\natexlab{}.
\newblock \showarticletitle{Disclosure of stuttering and quality of life in
  people who stutter}.
\newblock \bibinfo{journal}{\emph{Journal of fluency disorders}}
  \bibinfo{volume}{58} (\bibinfo{year}{2018}), \bibinfo{pages}{1--10}.
\newblock


\bibitem[\protect\citeauthoryear{Chee, Ai, and Yaacob}{Chee
  et~al\mbox{.}}{2009}]%
        {chee2009overview}
\bibfield{author}{\bibinfo{person}{Lim~Sin Chee}, \bibinfo{person}{Ooi~Chia
  Ai}, {and} \bibinfo{person}{Sazali Yaacob}.} \bibinfo{year}{2009}\natexlab{}.
\newblock \showarticletitle{Overview of automatic stuttering recognition
  system}. In \bibinfo{booktitle}{\emph{Proc. International Conference on
  Man-Machine Systems, no. October, Batu Ferringhi, Penang Malaysia}}.
  \bibinfo{pages}{1--6}.
\newblock


\bibitem[\protect\citeauthoryear{Chen, Huang, Hsieh, Kao, and Chang}{Chen
  et~al\mbox{.}}{2012}]%
        {chen2012flow}
\bibfield{author}{\bibinfo{person}{Mei-Hua Chen}, \bibinfo{person}{Shih-Ting
  Huang}, \bibinfo{person}{Hung-Ting Hsieh}, \bibinfo{person}{Ting-Hui Kao},
  {and} \bibinfo{person}{Jason~S Chang}.} \bibinfo{year}{2012}\natexlab{}.
\newblock \showarticletitle{FLOW: a first-language-oriented writing assistant
  system}. In \bibinfo{booktitle}{\emph{Proceedings of the ACL 2012 System
  Demonstrations}}. \bibinfo{pages}{157--162}.
\newblock


\bibitem[\protect\citeauthoryear{Chen, Lasko, Mei, Denny, and Xu}{Chen
  et~al\mbox{.}}{2015}]%
        {chen2015study}
\bibfield{author}{\bibinfo{person}{Yukun Chen}, \bibinfo{person}{Thomas~A
  Lasko}, \bibinfo{person}{Qiaozhu Mei}, \bibinfo{person}{Joshua~C Denny},
  {and} \bibinfo{person}{Hua Xu}.} \bibinfo{year}{2015}\natexlab{}.
\newblock \showarticletitle{A study of active learning methods for named entity
  recognition in clinical text}.
\newblock \bibinfo{journal}{\emph{Journal of biomedical informatics}}
  \bibinfo{volume}{58} (\bibinfo{year}{2015}), \bibinfo{pages}{11--18}.
\newblock


\bibitem[\protect\citeauthoryear{Constantino, Manning, and
  Nordstrom}{Constantino et~al\mbox{.}}{2017}]%
        {constantino2017rethinking}
\bibfield{author}{\bibinfo{person}{Christopher~Dominick Constantino},
  \bibinfo{person}{Walter~H Manning}, {and} \bibinfo{person}{Susan~Naomi
  Nordstrom}.} \bibinfo{year}{2017}\natexlab{}.
\newblock \showarticletitle{Rethinking covert stuttering}.
\newblock \bibinfo{journal}{\emph{Journal of Fluency Disorders}}
  \bibinfo{volume}{53} (\bibinfo{year}{2017}), \bibinfo{pages}{26--40}.
\newblock


\bibitem[\protect\citeauthoryear{Das, Mock, Chacon, Irani, Golob, and
  Najafirad}{Das et~al\mbox{.}}{2020}]%
        {das2020stuttering}
\bibfield{author}{\bibinfo{person}{Arun Das}, \bibinfo{person}{Jeffrey Mock},
  \bibinfo{person}{Henry Chacon}, \bibinfo{person}{Farzan Irani},
  \bibinfo{person}{Edward Golob}, {and} \bibinfo{person}{Peyman Najafirad}.}
  \bibinfo{year}{2020}\natexlab{}.
\newblock \showarticletitle{Stuttering Speech Disfluency Prediction using
  Explainable Attribution Vectors of Facial Muscle Movements}.
\newblock \bibinfo{journal}{\emph{arXiv preprint arXiv:2010.01231}}
  (\bibinfo{year}{2020}).
\newblock


\bibitem[\protect\citeauthoryear{Dasgupta and Hsu}{Dasgupta and Hsu}{2008}]%
        {dasgupta2008hierarchical}
\bibfield{author}{\bibinfo{person}{Sanjoy Dasgupta} {and}
  \bibinfo{person}{Daniel Hsu}.} \bibinfo{year}{2008}\natexlab{}.
\newblock \showarticletitle{Hierarchical sampling for active learning}. In
  \bibinfo{booktitle}{\emph{Proceedings of the 25th international conference on
  Machine learning}}. ACM, \bibinfo{pages}{208--215}.
\newblock


\bibitem[\protect\citeauthoryear{Devlin, Chang, Lee, and Toutanova}{Devlin
  et~al\mbox{.}}{2018}]%
        {bert}
\bibfield{author}{\bibinfo{person}{Jacob Devlin}, \bibinfo{person}{Ming-Wei
  Chang}, \bibinfo{person}{Kenton Lee}, {and} \bibinfo{person}{Kristina
  Toutanova}.} \bibinfo{year}{2018}\natexlab{}.
\newblock \showarticletitle{Bert: Pre-training of deep bidirectional
  transformers for language understanding}.
\newblock \bibinfo{journal}{\emph{arXiv preprint arXiv:1810.04805}}
  (\bibinfo{year}{2018}).
\newblock


\bibitem[\protect\citeauthoryear{Douglass, Kennedy, and Smith}{Douglass
  et~al\mbox{.}}{2020}]%
        {douglass2020speech}
\bibfield{author}{\bibinfo{person}{Jill~E Douglass}, \bibinfo{person}{Catherine
  Kennedy}, {and} \bibinfo{person}{Kaitlyn Smith}.}
  \bibinfo{year}{2020}\natexlab{}.
\newblock \showarticletitle{Speech-Language Therapy Experiences Across the Life
  Span of an Individual Who Covertly Stutters: A Thematic Analysis}.
\newblock \bibinfo{journal}{\emph{Perspectives of the ASHA Special Interest
  Groups}} \bibinfo{volume}{5}, \bibinfo{number}{6} (\bibinfo{year}{2020}),
  \bibinfo{pages}{1441--1453}.
\newblock


\bibitem[\protect\citeauthoryear{Esmaili, Dabanloo, and Vali}{Esmaili
  et~al\mbox{.}}{2017}]%
        {esmaili2017automatic}
\bibfield{author}{\bibinfo{person}{Iman Esmaili},
  \bibinfo{person}{Nader~Jafarnia Dabanloo}, {and} \bibinfo{person}{Mansour
  Vali}.} \bibinfo{year}{2017}\natexlab{}.
\newblock \showarticletitle{An automatic prolongation detection approach in
  continuous speech with robustness against speaking rate variations}.
\newblock \bibinfo{journal}{\emph{Journal of medical signals and sensors}}
  \bibinfo{volume}{7}, \bibinfo{number}{1} (\bibinfo{year}{2017}),
  \bibinfo{pages}{1}.
\newblock


\bibitem[\protect\citeauthoryear{Fassetti, Fassetti, and Nistic{\`o}}{Fassetti
  et~al\mbox{.}}{2019}]%
        {fassetti2019learning}
\bibfield{author}{\bibinfo{person}{Fabio Fassetti}, \bibinfo{person}{Ilaria
  Fassetti}, {and} \bibinfo{person}{Simona Nistic{\`o}}.}
  \bibinfo{year}{2019}\natexlab{}.
\newblock \showarticletitle{Learning and detecting stuttering disorders}. In
  \bibinfo{booktitle}{\emph{IFIP International Conference on Artificial
  Intelligence Applications and Innovations}}. Springer,
  \bibinfo{pages}{319--330}.
\newblock


\bibitem[\protect\citeauthoryear{Freund, Seung, Shamir, and Tishby}{Freund
  et~al\mbox{.}}{1997}]%
        {qbc2}
\bibfield{author}{\bibinfo{person}{Yoav Freund}, \bibinfo{person}{H~Sebastian
  Seung}, \bibinfo{person}{Eli Shamir}, {and} \bibinfo{person}{Naftali
  Tishby}.} \bibinfo{year}{1997}\natexlab{}.
\newblock \showarticletitle{Selective sampling using the query by committee
  algorithm}.
\newblock \bibinfo{journal}{\emph{Machine learning}} \bibinfo{volume}{28},
  \bibinfo{number}{2-3} (\bibinfo{year}{1997}), \bibinfo{pages}{133--168}.
\newblock


\bibitem[\protect\citeauthoryear{Frich, MacDonald~Vermeulen, Remy, Biskjaer,
  and Dalsgaard}{Frich et~al\mbox{.}}{2019}]%
        {frich2019mapping}
\bibfield{author}{\bibinfo{person}{Jonas Frich}, \bibinfo{person}{Lindsay
  MacDonald~Vermeulen}, \bibinfo{person}{Christian Remy},
  \bibinfo{person}{Michael~Mose Biskjaer}, {and} \bibinfo{person}{Peter
  Dalsgaard}.} \bibinfo{year}{2019}\natexlab{}.
\newblock \showarticletitle{Mapping the landscape of creativity support tools
  in HCI}. In \bibinfo{booktitle}{\emph{Proceedings of the 2019 CHI Conference
  on Human Factors in Computing Systems}}. \bibinfo{pages}{1--18}.
\newblock


\bibitem[\protect\citeauthoryear{Garcia-Barrera and Davidow}{Garcia-Barrera and
  Davidow}{2015}]%
        {anticipation}
\bibfield{author}{\bibinfo{person}{Mauricio~A Garcia-Barrera} {and}
  \bibinfo{person}{Jason~H Davidow}.} \bibinfo{year}{2015}\natexlab{}.
\newblock \showarticletitle{Anticipation in stuttering: A theoretical model of
  the nature of stutter prediction}.
\newblock \bibinfo{journal}{\emph{Journal of fluency disorders}}
  \bibinfo{volume}{44} (\bibinfo{year}{2015}), \bibinfo{pages}{1--15}.
\newblock


\bibitem[\protect\citeauthoryear{Ghai, Liao, Zhang, Bellamy, and Mueller}{Ghai
  et~al\mbox{.}}{2021}]%
        {xal}
\bibfield{author}{\bibinfo{person}{Bhavya Ghai}, \bibinfo{person}{Q.~Vera
  Liao}, \bibinfo{person}{Yunfeng Zhang}, \bibinfo{person}{Rachel Bellamy},
  {and} \bibinfo{person}{Klaus Mueller}.} \bibinfo{year}{2021}\natexlab{}.
\newblock \showarticletitle{Explainable Active Learning (XAL): Toward AI
  Explanations as Interfaces for Machine Teachers}.
\newblock \bibinfo{journal}{\emph{Proc. ACM Hum.-Comput. Interact.}}
  \bibinfo{volume}{4}, \bibinfo{number}{CSCW3}, Article
  \bibinfo{articleno}{235} (\bibinfo{date}{Jan.} \bibinfo{year}{2021}),
  \bibinfo{numpages}{28}~pages.
\newblock
\urldef\tempurl%
\url{https://doi.org/10.1145/3432934}
\showDOI{\tempurl}


\bibitem[\protect\citeauthoryear{Guitar}{Guitar}{2013}]%
        {guitar2013stuttering}
\bibfield{author}{\bibinfo{person}{Barry Guitar}.}
  \bibinfo{year}{2013}\natexlab{}.
\newblock \bibinfo{booktitle}{\emph{Stuttering: An integrated approach to its
  nature and treatment}}.
\newblock \bibinfo{publisher}{Lippincott Williams \& Wilkins}.
\newblock


\bibitem[\protect\citeauthoryear{Hanakovi{\v{c}} and Nagy}{Hanakovi{\v{c}} and
  Nagy}{2006}]%
        {hanakovivc2006speech}
\bibfield{author}{\bibinfo{person}{Tom{\'a}{\v{s}} Hanakovi{\v{c}}} {and}
  \bibinfo{person}{Marek Nagy}.} \bibinfo{year}{2006}\natexlab{}.
\newblock \showarticletitle{Speech recognition helps visually impaired people
  writing mathematical formulas}. In \bibinfo{booktitle}{\emph{International
  Conference on Computers for Handicapped Persons}}. Springer,
  \bibinfo{pages}{1231--1234}.
\newblock


\bibitem[\protect\citeauthoryear{Herder, Howard, Nye, and Vanryckeghem}{Herder
  et~al\mbox{.}}{2006}]%
        {herder2006effectiveness}
\bibfield{author}{\bibinfo{person}{Carl Herder}, \bibinfo{person}{Courtney
  Howard}, \bibinfo{person}{Chad Nye}, {and} \bibinfo{person}{Martine
  Vanryckeghem}.} \bibinfo{year}{2006}\natexlab{}.
\newblock \showarticletitle{Effectiveness of behavioral stuttering treatment: a
  systemic review and meta-analysis}.
\newblock \bibinfo{journal}{\emph{Contemporary Issues in Communication Science
  and Disorders}} \bibinfo{volume}{33}, \bibinfo{number}{Spring}
  (\bibinfo{year}{2006}), \bibinfo{pages}{61--73}.
\newblock


\bibitem[\protect\citeauthoryear{Howell, Au-Yeung, Yaruss, and Eldridge}{Howell
  et~al\mbox{.}}{2006}]%
        {howell2006phonetic}
\bibfield{author}{\bibinfo{person}{Peter Howell}, \bibinfo{person}{James
  Au-Yeung}, \bibinfo{person}{Scott~J Yaruss}, {and} \bibinfo{person}{Kevin
  Eldridge}.} \bibinfo{year}{2006}\natexlab{}.
\newblock \showarticletitle{Phonetic difficulty and stuttering in English}.
\newblock \bibinfo{journal}{\emph{Clinical linguistics \& phonetics}}
  \bibinfo{volume}{20}, \bibinfo{number}{9} (\bibinfo{year}{2006}),
  \bibinfo{pages}{703--716}.
\newblock


\bibitem[\protect\citeauthoryear{Huang, Jin, and Zhou}{Huang
  et~al\mbox{.}}{2010}]%
        {quire}
\bibfield{author}{\bibinfo{person}{Sheng-Jun Huang}, \bibinfo{person}{Rong
  Jin}, {and} \bibinfo{person}{Zhi-Hua Zhou}.} \bibinfo{year}{2010}\natexlab{}.
\newblock \showarticletitle{Active learning by querying informative and
  representative examples}. In \bibinfo{booktitle}{\emph{Advances in neural
  information processing systems}}. \bibinfo{pages}{892--900}.
\newblock


\bibitem[\protect\citeauthoryear{Hulit and Wirtz}{Hulit and Wirtz}{1994}]%
        {hulit1994association}
\bibfield{author}{\bibinfo{person}{Lloyd~M Hulit} {and}
  \bibinfo{person}{Lauralee Wirtz}.} \bibinfo{year}{1994}\natexlab{}.
\newblock \showarticletitle{The association of attitudes toward stuttering with
  selected variables}.
\newblock \bibinfo{journal}{\emph{Journal of fluency disorders}}
  \bibinfo{volume}{19}, \bibinfo{number}{4} (\bibinfo{year}{1994}),
  \bibinfo{pages}{247--267}.
\newblock


\bibitem[\protect\citeauthoryear{Ivanova and Eriksen}{Ivanova and
  Eriksen}{2012}]%
        {ivanova2012bilingual}
\bibfield{author}{\bibinfo{person}{Nedelina Ivanova} {and}
  \bibinfo{person}{Olle Eriksen}.} \bibinfo{year}{2012}\natexlab{}.
\newblock \showarticletitle{A Bilingual Bimodal Reading And Writing Tool For
  Sign Language Users}.
\newblock \bibinfo{journal}{\emph{Universal Learning Design, Linz 2012}}
  (\bibinfo{year}{2012}), \bibinfo{pages}{79}.
\newblock


\bibitem[\protect\citeauthoryear{Jiang, Lu, Peng, Zhu, and Howell}{Jiang
  et~al\mbox{.}}{2012}]%
        {brain_activity}
\bibfield{author}{\bibinfo{person}{Jing Jiang}, \bibinfo{person}{Chunming Lu},
  \bibinfo{person}{Danling Peng}, \bibinfo{person}{Chaozhe Zhu}, {and}
  \bibinfo{person}{Peter Howell}.} \bibinfo{year}{2012}\natexlab{}.
\newblock \showarticletitle{Classification of types of stuttering symptoms
  based on brain activity}.
\newblock \bibinfo{journal}{\emph{PloS one}} \bibinfo{volume}{7},
  \bibinfo{number}{6} (\bibinfo{year}{2012}), \bibinfo{pages}{e39747}.
\newblock


\bibitem[\protect\citeauthoryear{Kagy, Kayadelen, Ma, Rostamizadeh, and
  Strnadova}{Kagy et~al\mbox{.}}{2019}]%
        {kagy2019practical}
\bibfield{author}{\bibinfo{person}{Jean-Fran{\c{c}}ois Kagy},
  \bibinfo{person}{Tolga Kayadelen}, \bibinfo{person}{Ji Ma},
  \bibinfo{person}{Afshin Rostamizadeh}, {and} \bibinfo{person}{Jana
  Strnadova}.} \bibinfo{year}{2019}\natexlab{}.
\newblock \showarticletitle{The Practical Challenges of Active Learning:
  Lessons Learned from Live Experimentation}.
\newblock \bibinfo{journal}{\emph{arXiv preprint arXiv:1907.00038}}
  (\bibinfo{year}{2019}).
\newblock


\bibitem[\protect\citeauthoryear{Kalinowski and Saltuklaroglu}{Kalinowski and
  Saltuklaroglu}{2005}]%
        {kalinowski2005stuttering}
\bibfield{author}{\bibinfo{person}{Joseph~S Kalinowski} {and}
  \bibinfo{person}{Tim Saltuklaroglu}.} \bibinfo{year}{2005}\natexlab{}.
\newblock \bibinfo{booktitle}{\emph{Stuttering}}.
\newblock \bibinfo{publisher}{Plural Publishing}.
\newblock


\bibitem[\protect\citeauthoryear{Karyuatry}{Karyuatry}{2018}]%
        {karyuatry2018grammarly}
\bibfield{author}{\bibinfo{person}{Laksnoria Karyuatry}.}
  \bibinfo{year}{2018}\natexlab{}.
\newblock \showarticletitle{Grammarly as a Tool to Improve Students’ Writing
  Quality: Free Online-Proofreader across the Boundaries}.
\newblock \bibinfo{journal}{\emph{JSSH (Jurnal Sains Sosial dan Humaniora)}}
  \bibinfo{volume}{2}, \bibinfo{number}{1} (\bibinfo{year}{2018}),
  \bibinfo{pages}{83--89}.
\newblock


\bibitem[\protect\citeauthoryear{Kinnunen, Leisma, Machunik, Kakkonen, and
  Lebrun}{Kinnunen et~al\mbox{.}}{2012}]%
        {SWAN}
\bibfield{author}{\bibinfo{person}{Tomi Kinnunen}, \bibinfo{person}{Henri
  Leisma}, \bibinfo{person}{Monika Machunik}, \bibinfo{person}{Tuomo Kakkonen},
  {and} \bibinfo{person}{Jean-Luc Lebrun}.} \bibinfo{year}{2012}\natexlab{}.
\newblock \showarticletitle{SWAN - Scientific Writing AssistaNt: A Tool for
  Helping Scholars to Write Reader-Friendly Manuscripts}. In
  \bibinfo{booktitle}{\emph{Proceedings of the Demonstrations at the 13th
  Conference of the European Chapter of the Association for Computational
  Linguistics}} (Avignon, France) \emph{(\bibinfo{series}{EACL '12})}.
  \bibinfo{publisher}{Association for Computational Linguistics},
  \bibinfo{address}{USA}, \bibinfo{pages}{20–24}.
\newblock


\bibitem[\protect\citeauthoryear{Klein and Hood}{Klein and Hood}{2004}]%
        {klein2004impact}
\bibfield{author}{\bibinfo{person}{Joseph~F Klein} {and}
  \bibinfo{person}{Stephen~B Hood}.} \bibinfo{year}{2004}\natexlab{}.
\newblock \showarticletitle{The impact of stuttering on employment
  opportunities and job performance}.
\newblock \bibinfo{journal}{\emph{Journal of fluency disorders}}
  \bibinfo{volume}{29}, \bibinfo{number}{4} (\bibinfo{year}{2004}),
  \bibinfo{pages}{255--273}.
\newblock


\bibitem[\protect\citeauthoryear{Kourkounakis, Hajavi, and Etemad}{Kourkounakis
  et~al\mbox{.}}{2020}]%
        {fluentnet}
\bibfield{author}{\bibinfo{person}{Tedd Kourkounakis},
  \bibinfo{person}{Amirhossein Hajavi}, {and} \bibinfo{person}{Ali Etemad}.}
  \bibinfo{year}{2020}\natexlab{}.
\newblock \showarticletitle{FluentNet: End-to-End Detection of Speech
  Disfluency with Deep Learning}.
\newblock \bibinfo{journal}{\emph{arXiv preprint arXiv:2009.11394}}
  (\bibinfo{year}{2020}).
\newblock


\bibitem[\protect\citeauthoryear{Kully and Boberg}{Kully and Boberg}{1988}]%
        {kully1988investigation}
\bibfield{author}{\bibinfo{person}{Deborah Kully} {and} \bibinfo{person}{Einer
  Boberg}.} \bibinfo{year}{1988}\natexlab{}.
\newblock \showarticletitle{An investigation of interclinic agreement in the
  identification of fluent and stuttered syllables}.
\newblock \bibinfo{journal}{\emph{Journal of fluency disorders}}
  \bibinfo{volume}{13}, \bibinfo{number}{5} (\bibinfo{year}{1988}),
  \bibinfo{pages}{309--318}.
\newblock


\bibitem[\protect\citeauthoryear{Lewis and Gale}{Lewis and Gale}{1994}]%
        {unc}
\bibfield{author}{\bibinfo{person}{David~D Lewis} {and}
  \bibinfo{person}{William~A Gale}.} \bibinfo{year}{1994}\natexlab{}.
\newblock \showarticletitle{A sequential algorithm for training text
  classifiers}. In \bibinfo{booktitle}{\emph{SIGIR’94}}. Springer,
  \bibinfo{pages}{3--12}.
\newblock


\bibitem[\protect\citeauthoryear{Lowell, Lipton, and Wallace}{Lowell
  et~al\mbox{.}}{2018}]%
        {lowell2018practical}
\bibfield{author}{\bibinfo{person}{David Lowell}, \bibinfo{person}{Zachary~C
  Lipton}, {and} \bibinfo{person}{Byron~C Wallace}.}
  \bibinfo{year}{2018}\natexlab{}.
\newblock \showarticletitle{Practical obstacles to deploying active learning}.
\newblock \bibinfo{journal}{\emph{arXiv preprint arXiv:1807.04801}}
  (\bibinfo{year}{2018}).
\newblock


\bibitem[\protect\citeauthoryear{Mikolov, Chen, Corrado, and Dean}{Mikolov
  et~al\mbox{.}}{2013}]%
        {word2vec}
\bibfield{author}{\bibinfo{person}{Tomas Mikolov}, \bibinfo{person}{Kai Chen},
  \bibinfo{person}{Greg Corrado}, {and} \bibinfo{person}{Jeffrey Dean}.}
  \bibinfo{year}{2013}\natexlab{}.
\newblock \showarticletitle{Efficient estimation of word representations in
  vector space}.
\newblock \bibinfo{journal}{\emph{arXiv preprint arXiv:1301.3781}}
  (\bibinfo{year}{2013}).
\newblock


\bibitem[\protect\citeauthoryear{Miller}{Miller}{1995}]%
        {wordnet}
\bibfield{author}{\bibinfo{person}{George~A Miller}.}
  \bibinfo{year}{1995}\natexlab{}.
\newblock \showarticletitle{WordNet: a lexical database for English}.
\newblock \bibinfo{journal}{\emph{Commun. ACM}} \bibinfo{volume}{38},
  \bibinfo{number}{11} (\bibinfo{year}{1995}), \bibinfo{pages}{39--41}.
\newblock


\bibitem[\protect\citeauthoryear{Murphy, Quesal, and Gulker}{Murphy
  et~al\mbox{.}}{2007}]%
        {murphy2007covert}
\bibfield{author}{\bibinfo{person}{Bill Murphy}, \bibinfo{person}{Robert~W
  Quesal}, {and} \bibinfo{person}{Hope Gulker}.}
  \bibinfo{year}{2007}\natexlab{}.
\newblock \showarticletitle{Covert stuttering}.
\newblock \bibinfo{journal}{\emph{Perspectives on Fluency and Fluency
  Disorders}} \bibinfo{volume}{17}, \bibinfo{number}{2} (\bibinfo{year}{2007}),
  \bibinfo{pages}{4--9}.
\newblock


\bibitem[\protect\citeauthoryear{Parrish}{Parrish}{2017}]%
        {parrish2017poetic}
\bibfield{author}{\bibinfo{person}{Allison Parrish}.}
  \bibinfo{year}{2017}\natexlab{}.
\newblock \showarticletitle{Poetic sound similarity vectors using phonetic
  features}. In \bibinfo{booktitle}{\emph{Thirteenth Artificial Intelligence
  and Interactive Digital Entertainment Conference}}.
\newblock


\bibitem[\protect\citeauthoryear{Plexico, Manning, and Levitt}{Plexico
  et~al\mbox{.}}{2009}]%
        {plexico2009coping}
\bibfield{author}{\bibinfo{person}{Laura~W Plexico}, \bibinfo{person}{Walter~H
  Manning}, {and} \bibinfo{person}{Heidi Levitt}.}
  \bibinfo{year}{2009}\natexlab{}.
\newblock \showarticletitle{Coping responses by adults who stutter: Part I.
  Protecting the self and others}.
\newblock \bibinfo{journal}{\emph{Journal of fluency disorders}}
  \bibinfo{volume}{34}, \bibinfo{number}{2} (\bibinfo{year}{2009}),
  \bibinfo{pages}{87--107}.
\newblock


\bibitem[\protect\citeauthoryear{Prasse and Kikano}{Prasse and Kikano}{2008}]%
        {prasse2008stuttering}
\bibfield{author}{\bibinfo{person}{Jane~E Prasse} {and}
  \bibinfo{person}{George~E Kikano}.} \bibinfo{year}{2008}\natexlab{}.
\newblock \showarticletitle{Stuttering: an overview}.
\newblock \bibinfo{journal}{\emph{American family physician}}
  \bibinfo{volume}{77}, \bibinfo{number}{9} (\bibinfo{year}{2008}),
  \bibinfo{pages}{1271--1276}.
\newblock


\bibitem[\protect\citeauthoryear{Ramirez-Loaiza, Sharma, Kumar, and
  Bilgic}{Ramirez-Loaiza et~al\mbox{.}}{2017}]%
        {ramirez2017active}
\bibfield{author}{\bibinfo{person}{Maria~E Ramirez-Loaiza},
  \bibinfo{person}{Manali Sharma}, \bibinfo{person}{Geet Kumar}, {and}
  \bibinfo{person}{Mustafa Bilgic}.} \bibinfo{year}{2017}\natexlab{}.
\newblock \showarticletitle{Active learning: an empirical study of common
  baselines}.
\newblock \bibinfo{journal}{\emph{Data mining and knowledge discovery}}
  \bibinfo{volume}{31}, \bibinfo{number}{2} (\bibinfo{year}{2017}),
  \bibinfo{pages}{287--313}.
\newblock


\bibitem[\protect\citeauthoryear{Ravikumar, Reddy, Rajagopal, and
  Nagaraj}{Ravikumar et~al\mbox{.}}{2008}]%
        {ravikumar2008automatic}
\bibfield{author}{\bibinfo{person}{KM Ravikumar}, \bibinfo{person}{Balakrishna
  Reddy}, \bibinfo{person}{R Rajagopal}, {and} \bibinfo{person}{H Nagaraj}.}
  \bibinfo{year}{2008}\natexlab{}.
\newblock \showarticletitle{Automatic detection of syllable repetition in read
  speech for objective assessment of stuttered disfluencies}.
\newblock \bibinfo{journal}{\emph{Proceedings of world academy science,
  engineering and technology}}  \bibinfo{volume}{36} (\bibinfo{year}{2008}),
  \bibinfo{pages}{270--273}.
\newblock


\bibitem[\protect\citeauthoryear{Roemmele and Gordon}{Roemmele and
  Gordon}{2015}]%
        {creative}
\bibfield{author}{\bibinfo{person}{Melissa Roemmele} {and}
  \bibinfo{person}{Andrew~S Gordon}.} \bibinfo{year}{2015}\natexlab{}.
\newblock \showarticletitle{Creative help: A story writing assistant}. In
  \bibinfo{booktitle}{\emph{International Conference on Interactive Digital
  Storytelling}}. Springer, \bibinfo{pages}{81--92}.
\newblock


\bibitem[\protect\citeauthoryear{Sanghrajka, Hidalgo, Chen, and
  Kapadia}{Sanghrajka et~al\mbox{.}}{2017}]%
        {sanghrajka2017lisa}
\bibfield{author}{\bibinfo{person}{Rushit Sanghrajka}, \bibinfo{person}{Daniel
  Hidalgo}, \bibinfo{person}{Patrick Chen}, {and} \bibinfo{person}{Mubbasir
  Kapadia}.} \bibinfo{year}{2017}\natexlab{}.
\newblock \showarticletitle{Lisa: Lexically intelligent story assistant}. In
  \bibinfo{booktitle}{\emph{Proceedings of the AAAI Conference on Artificial
  Intelligence and Interactive Digital Entertainment}},
  Vol.~\bibinfo{volume}{13}.
\newblock


\bibitem[\protect\citeauthoryear{Settles}{Settles}{2009}]%
        {settles2009active}
\bibfield{author}{\bibinfo{person}{Burr Settles}.}
  \bibinfo{year}{2009}\natexlab{}.
\newblock \bibinfo{booktitle}{\emph{Active learning literature survey}}.
\newblock \bibinfo{type}{{T}echnical {R}eport}.
  \bibinfo{institution}{University of Wisconsin-Madison Department of Computer
  Sciences}.
\newblock


\bibitem[\protect\citeauthoryear{Settles}{Settles}{2011}]%
        {settles2011theories}
\bibfield{author}{\bibinfo{person}{Burr Settles}.}
  \bibinfo{year}{2011}\natexlab{}.
\newblock \showarticletitle{From theories to queries: Active learning in
  practice}. In \bibinfo{booktitle}{\emph{Active Learning and Experimental
  Design workshop In conjunction with AISTATS 2010}}. JMLR Workshop and
  Conference Proceedings, \bibinfo{pages}{1--18}.
\newblock


\bibitem[\protect\citeauthoryear{Seung, Opper, and Sompolinsky}{Seung
  et~al\mbox{.}}{1992}]%
        {qbc}
\bibfield{author}{\bibinfo{person}{H~Sebastian Seung}, \bibinfo{person}{Manfred
  Opper}, {and} \bibinfo{person}{Haim Sompolinsky}.}
  \bibinfo{year}{1992}\natexlab{}.
\newblock \showarticletitle{Query by committee}. In
  \bibinfo{booktitle}{\emph{Proceedings of the fifth annual workshop on
  Computational learning theory}}. ACM, \bibinfo{pages}{287--294}.
\newblock


\bibitem[\protect\citeauthoryear{Silverman and Bongey}{Silverman and
  Bongey}{1997}]%
        {silverman1997nurses}
\bibfield{author}{\bibinfo{person}{Franklin~H Silverman} {and}
  \bibinfo{person}{Tamara~A Bongey}.} \bibinfo{year}{1997}\natexlab{}.
\newblock \showarticletitle{Nurses' attitudes toward physicians who stutter}.
\newblock \bibinfo{journal}{\emph{Journal of Fluency Disorders}}
  \bibinfo{volume}{1}, \bibinfo{number}{22} (\bibinfo{year}{1997}),
  \bibinfo{pages}{61--62}.
\newblock


\bibitem[\protect\citeauthoryear{Silverman and Paynter}{Silverman and
  Paynter}{1990}]%
        {silverman1990impact}
\bibfield{author}{\bibinfo{person}{Franklin~H Silverman} {and}
  \bibinfo{person}{Kathryn~K Paynter}.} \bibinfo{year}{1990}\natexlab{}.
\newblock \showarticletitle{Impact of stuttering on perception of occupational
  competence}.
\newblock \bibinfo{journal}{\emph{Journal of Fluency Disorders}}
  \bibinfo{volume}{15}, \bibinfo{number}{2} (\bibinfo{year}{1990}),
  \bibinfo{pages}{87--91}.
\newblock


\bibitem[\protect\citeauthoryear{Song, Wang, Gao, and An}{Song
  et~al\mbox{.}}{2018}]%
        {song2018active}
\bibfield{author}{\bibinfo{person}{Jinhua Song}, \bibinfo{person}{Hao Wang},
  \bibinfo{person}{Yang Gao}, {and} \bibinfo{person}{Bo An}.}
  \bibinfo{year}{2018}\natexlab{}.
\newblock \showarticletitle{Active learning with confidence-based answers for
  crowdsourcing labeling tasks}.
\newblock \bibinfo{journal}{\emph{Knowledge-Based Systems}}
  \bibinfo{volume}{159} (\bibinfo{year}{2018}), \bibinfo{pages}{244--258}.
\newblock


\bibitem[\protect\citeauthoryear{Starkweather}{Starkweather}{1987}]%
        {starkweather1987fluency}
\bibfield{author}{\bibinfo{person}{C~Woodruff Starkweather}.}
  \bibinfo{year}{1987}\natexlab{}.
\newblock \bibinfo{booktitle}{\emph{Fluency and stuttering.}}
\newblock \bibinfo{publisher}{Prentice-Hall, Inc}.
\newblock


\bibitem[\protect\citeauthoryear{{\'S}wietlicka, Kuniszyk-J{\'o}{\'z}kowiak,
  and Smo{\l}ka}{{\'S}wietlicka et~al\mbox{.}}{2013}]%
        {swietlicka2013hierarchical}
\bibfield{author}{\bibinfo{person}{Izabela {\'S}wietlicka},
  \bibinfo{person}{Wies{\l}awa Kuniszyk-J{\'o}{\'z}kowiak}, {and}
  \bibinfo{person}{El{\.z}bieta Smo{\l}ka}.} \bibinfo{year}{2013}\natexlab{}.
\newblock \showarticletitle{Hierarchical ANN system for stuttering
  identification}.
\newblock \bibinfo{journal}{\emph{Computer Speech \& Language}}
  \bibinfo{volume}{27}, \bibinfo{number}{1} (\bibinfo{year}{2013}),
  \bibinfo{pages}{228--242}.
\newblock


\bibitem[\protect\citeauthoryear{Textio}{Textio}{line}]%
        {textio}
\bibfield{author}{\bibinfo{person}{Textio}.}
  \bibinfo{year}{[Online]}\natexlab{}.
\newblock \bibinfo{title}{{The augmented writing platform}}.
\newblock
\newblock
\newblock
\shownote{\url{https://textio.com/} (accessed 4 April 2021).}


\bibitem[\protect\citeauthoryear{Tichenor and Yaruss}{Tichenor and
  Yaruss}{2019}]%
        {tichenor2019group}
\bibfield{author}{\bibinfo{person}{Seth~E Tichenor} {and}
  \bibinfo{person}{J~Scott Yaruss}.} \bibinfo{year}{2019}\natexlab{}.
\newblock \showarticletitle{Group experiences and individual differences in
  stuttering}.
\newblock \bibinfo{journal}{\emph{Journal of Speech, Language, and Hearing
  Research}} \bibinfo{volume}{62}, \bibinfo{number}{12} (\bibinfo{year}{2019}),
  \bibinfo{pages}{4335--4350}.
\newblock


\bibitem[\protect\citeauthoryear{Tichenor and Yaruss}{Tichenor and
  Yaruss}{2021}]%
        {variability}
\bibfield{author}{\bibinfo{person}{Seth~E Tichenor} {and}
  \bibinfo{person}{J~Scott Yaruss}.} \bibinfo{year}{2021}\natexlab{}.
\newblock \showarticletitle{Variability of stuttering: Behavior and impact}.
\newblock \bibinfo{journal}{\emph{American Journal of Speech-Language
  Pathology}} \bibinfo{volume}{30}, \bibinfo{number}{1} (\bibinfo{year}{2021}),
  \bibinfo{pages}{75--88}.
\newblock


\bibitem[\protect\citeauthoryear{Tong and Koller}{Tong and Koller}{2001}]%
        {tong2001support}
\bibfield{author}{\bibinfo{person}{Simon Tong} {and} \bibinfo{person}{Daphne
  Koller}.} \bibinfo{year}{2001}\natexlab{}.
\newblock \showarticletitle{Support vector machine active learning with
  applications to text classification}.
\newblock \bibinfo{journal}{\emph{Journal of machine learning research}}
  \bibinfo{volume}{2}, \bibinfo{number}{Nov} (\bibinfo{year}{2001}),
  \bibinfo{pages}{45--66}.
\newblock


\bibitem[\protect\citeauthoryear{van Staden and Kamper}{van Staden and
  Kamper}{2020}]%
        {van2020improving}
\bibfield{author}{\bibinfo{person}{Lisa van Staden} {and}
  \bibinfo{person}{Herman Kamper}.} \bibinfo{year}{2020}\natexlab{}.
\newblock \showarticletitle{Improving unsupervised acoustic word embeddings
  using speaker and gender information}. In \bibinfo{booktitle}{\emph{2020
  International SAUPEC/RobMech/PRASA Conference}}. IEEE, \bibinfo{pages}{1--6}.
\newblock


\bibitem[\protect\citeauthoryear{Verdon, Wilson, Smith-Tamaray, and
  McAllister}{Verdon et~al\mbox{.}}{2011}]%
        {verdon2011investigation}
\bibfield{author}{\bibinfo{person}{Sarah Verdon}, \bibinfo{person}{Linda
  Wilson}, \bibinfo{person}{Michelle Smith-Tamaray}, {and}
  \bibinfo{person}{Lindy McAllister}.} \bibinfo{year}{2011}\natexlab{}.
\newblock \showarticletitle{An investigation of equity of rural speech-language
  pathology services for children: A geographic perspective}.
\newblock \bibinfo{journal}{\emph{International Journal of Speech-Language
  Pathology}} \bibinfo{volume}{13}, \bibinfo{number}{3} (\bibinfo{year}{2011}),
  \bibinfo{pages}{239--250}.
\newblock


\bibitem[\protect\citeauthoryear{Villegas, Flores, Acu{\~n}a, Pacheco-Barrios,
  and Elias}{Villegas et~al\mbox{.}}{2019}]%
        {villegas2019novel}
\bibfield{author}{\bibinfo{person}{Bruno Villegas}, \bibinfo{person}{Kevin~M
  Flores}, \bibinfo{person}{Kevin~Jos{\'e} Acu{\~n}a}, \bibinfo{person}{Kevin
  Pacheco-Barrios}, {and} \bibinfo{person}{Dante Elias}.}
  \bibinfo{year}{2019}\natexlab{}.
\newblock \showarticletitle{A novel stuttering disfluency classification system
  based on respiratory biosignals}. In \bibinfo{booktitle}{\emph{2019 41st
  Annual International Conference of the IEEE Engineering in Medicine and
  Biology Society (EMBC)}}. IEEE, \bibinfo{pages}{4660--4663}.
\newblock


\bibitem[\protect\citeauthoryear{Waqar, Aslam, and Farhan}{Waqar
  et~al\mbox{.}}{2019}]%
        {waqar2019intelligent}
\bibfield{author}{\bibinfo{person}{Mirza~Muhammad Waqar},
  \bibinfo{person}{Muhammad Aslam}, {and} \bibinfo{person}{Muhammad Farhan}.}
  \bibinfo{year}{2019}\natexlab{}.
\newblock \showarticletitle{An intelligent and interactive interface to support
  symmetrical collaborative educational writing among visually impaired and
  sighted users}.
\newblock \bibinfo{journal}{\emph{Symmetry}} \bibinfo{volume}{11},
  \bibinfo{number}{2} (\bibinfo{year}{2019}), \bibinfo{pages}{238}.
\newblock


\bibitem[\protect\citeauthoryear{Wright, Shaikh, Park, Epperson, Ahmed, Pinel,
  Chau, and Yang}{Wright et~al\mbox{.}}{2021}]%
        {wright2021recast}
\bibfield{author}{\bibinfo{person}{Austin~P Wright}, \bibinfo{person}{Omar
  Shaikh}, \bibinfo{person}{Haekyu Park}, \bibinfo{person}{Will Epperson},
  \bibinfo{person}{Muhammed Ahmed}, \bibinfo{person}{Stephane Pinel},
  \bibinfo{person}{Duen~Horng Chau}, {and} \bibinfo{person}{Diyi Yang}.}
  \bibinfo{year}{2021}\natexlab{}.
\newblock \showarticletitle{RECAST: Enabling User Recourse and Interpretability
  of Toxicity Detection Models with Interactive Visualization}.
\newblock \bibinfo{journal}{\emph{Proceedings of the ACM on Human-Computer
  Interaction}} \bibinfo{volume}{5}, \bibinfo{number}{CSCW1}
  (\bibinfo{year}{2021}), \bibinfo{pages}{1--26}.
\newblock


\bibitem[\protect\citeauthoryear{Wu, Reynolds, Li, and Guzm{\'a}n}{Wu
  et~al\mbox{.}}{2019}]%
        {wu2019design}
\bibfield{author}{\bibinfo{person}{Shaomei Wu}, \bibinfo{person}{Lindsay
  Reynolds}, \bibinfo{person}{Xian Li}, {and} \bibinfo{person}{Francisco
  Guzm{\'a}n}.} \bibinfo{year}{2019}\natexlab{}.
\newblock \showarticletitle{Design and evaluation of a social media writing
  support tool for people with dyslexia}. In
  \bibinfo{booktitle}{\emph{Proceedings of the 2019 CHI Conference on Human
  Factors in Computing Systems}}. \bibinfo{pages}{1--14}.
\newblock


\bibitem[\protect\citeauthoryear{Yaruss}{Yaruss}{2010}]%
        {yaruss2010assessing}
\bibfield{author}{\bibinfo{person}{J~Scott Yaruss}.}
  \bibinfo{year}{2010}\natexlab{}.
\newblock \showarticletitle{Assessing quality of life in stuttering treatment
  outcomes research}.
\newblock \bibinfo{journal}{\emph{Journal of fluency disorders}}
  \bibinfo{volume}{35}, \bibinfo{number}{3} (\bibinfo{year}{2010}),
  \bibinfo{pages}{190--202}.
\newblock


\bibitem[\protect\citeauthoryear{Zhu}{Zhu}{2005}]%
        {zhu2005semi}
\bibfield{author}{\bibinfo{person}{Xiaojin~Jerry Zhu}.}
  \bibinfo{year}{2005}\natexlab{}.
\newblock \showarticletitle{Semi-supervised learning literature survey}.
\newblock  (\bibinfo{year}{2005}).
\newblock


\bibitem[\protect\citeauthoryear{Zimmermann, Liljeblad, Frank, and
  Cleeland}{Zimmermann et~al\mbox{.}}{1983}]%
        {zimmermann1983indians}
\bibfield{author}{\bibinfo{person}{Gerald Zimmermann}, \bibinfo{person}{Sven
  Liljeblad}, \bibinfo{person}{Arthur Frank}, {and} \bibinfo{person}{Charlotte
  Cleeland}.} \bibinfo{year}{1983}\natexlab{}.
\newblock \showarticletitle{The Indians have many terms for it: Stuttering
  among the Bannock-Shoshoni}.
\newblock \bibinfo{journal}{\emph{Journal of Speech, Language, and Hearing
  Research}} \bibinfo{volume}{26}, \bibinfo{number}{2} (\bibinfo{year}{1983}),
  \bibinfo{pages}{315--318}.
\newblock


\end{thebibliography}

\end{document}


\title{APPENDIX}
\maketitle
\section{User Profiles}
We wanted to evaluate our system on a diverse set of user profiles with varying degrees of stuttering. Moreover, we also wanted our user profiles to mimic reality.  
Our paper presents 10 different user profiles. One of them is derived from the personal experiences of an author of this paper who stutters. The next 4 user profiles are based on self reported data from stuttering communities on facebook and reddit. The last 5 profiles are a combination of two or three phonetic patterns from the top 5 profiles. This was intended to create more complex cases to test our system.
The following table presents the underlying phonetic pattern and seed words for all 10 users. Users 6-10 have more serious stuttering issues as they stutter on multiple patterns. For example, User 6 stutter on all words starting with a consonant and then followed by the ‘r’ sound (pattern 1) as well as words starting with B,P,D,M,N, and F (pattern 5). 

\begin{table}[h!]
    \centering
    \begin{tabular}{ p{0.08\linewidth} | p{0.31\linewidth} | p{0.25\linewidth} | p{0.25\linewidth}} 
 \hline
 User ID & Phonetic Pattern & Easy words & Difficult Words \\
 \hline
 User 1	& words starting with a consonant and then followed by the ‘r’ sound & clock, regular, water, made, computer	& graph, group, green, grand, grapes \\
 \hline
 User 2	& words starting with ‘st’ or ‘fl’ & the, cat, owl, bat, kite	& street, florida, straight, stutter, flexible \\
 \hline
User 3 & words with ‘r’ or ‘l’ letter at second place & about, people, day, other, kiwi & crisp, crumble, alaska, close, brisk \\
\hline
User 4 & words with ‘ch’ or ‘sc’ anywhere in the word & book, table, cat, shirt, window & scold, chair, beach, chase, fiscal \\
\hline
User 5 & words starting with B,P,D,M,N, and F & horse, house, group, actor, echo & packet, more, nostalgia, fish, boat \\
\hline
User 6 & pattern 1 + pattern 5 & racket, choice, egg, active, card & crime, provost, post, dragon, basket \\
\hline
User 7 & pattern 2 + pattern 4 & packet, more, nostalgia, fish, mouse & flood, scandal, stay, choke, discard \\
\hline
User 8 & pattern 3 + pattern 5 & cook, table, cat, she, jacket & graph, alcohol, ball, market, fancy \\
\hline
User 9 & pattern 1+ pattern 4+ pattern 5 & rational, recommend, circle, gang, tie & scam, grand, match, cheese, nose \\
\hline
User 10 & pattern 2 + pattern 3 + pattern 4 & beauty, pen, dish, govern, wire & match, scam, alcohol, scold, strict \\
 \hline
\end{tabular}
\end{table}